\documentclass[11pt,a4paper]{article}
\usepackage[hyperref]{acl2018}
\usepackage{times}
\usepackage{latexsym}
\usepackage{float}

\usepackage{url}
\usepackage{csquotes}
\usepackage{graphicx}
\newtheorem{exmp}{Example}[section]
\usepackage[bottom]{footmisc}
\usepackage{hyperref}
\usepackage{amsmath}
\usepackage{tabularx}
\usepackage[T1]{fontenc}
\usepackage{makecell}

\aclfinalcopy % Uncomment this line for the final submission
\def\aclpaperid{67} %  Enter the acl Paper ID here

\title{Relation extraction between the clinical entities based on the shortest dependency path based LSTM}

\author{Dhanachandra Ningthoujam$^{\dagger}$, Shweta Yadav$^{\ast}$, Pushpak Bhattacharyya$^{\ast}$, Asif Ekbal$^{\ast}$ \\
  $^{\ast}$Indian Institute of Technology Patna, India  \\
 $^{\dagger}$ezDI, INC\\
  {\tt $^{\ast}$\{shweta.pcs14,pb,asif\}@{iitp.ac.in}, $^{\dagger}$dhanachandra.n@ezdi.com}}

\date{}

\begin{document}
\maketitle
\begin{abstract}
Owing to the exponential rise in the electronic medical records, information extraction in this domain is becoming an important area of research in recent years. Relation extraction between the medical concepts such as medical problem, treatment, and test etc. is also one of the most important tasks in this area. In this paper, we present an efficient relation extraction system based on the shortest dependency path (SDP) generated from the dependency parsed tree of the sentence. Instead of relying on many handcrafted features and the whole sequence of tokens present in a sentence, our system relies only on the SDP between the target entities. For every pair of entities, the system takes only the words in the SDP, their dependency labels, Part-of-Speech information and the types of the entities as the input. We develop a dependency parser for extracting dependency information. We perform our experiments on the benchmark i2b2 dataset for clinical relation extraction challenge 2010. Experimental results show that our system outperforms the existing systems. %Our experiment also shows that the performance of our proposed parser is very close to the Standford parser in the relation extraction. 
\end{abstract}
\section{Introduction}
In recent years, the amount of clinical texts in electronic format has drastically increased. 
However, most of this information such as clinical named entities, the relationship between the clinical entities, medical summary etc. are still embedded in the form of unstructured text. Information are mostly available in an unstructured format. There is a necessity to investigate proper methods for making these information structured so that relevant information can be extracted at ease. %and The availability mainly in nonstandard natural language makes it difficult to automatically collect and present this information in a structured way. 
Automatic extraction of this information is very essential for the other applications such as clinical decision-making, clinical trial screening, and pharmacovigilance. As a result, the information extraction in this domain is becoming an interesting area of research.\\
Relation extraction is one of the important tasks in Natural Language Processing (NLP) that aims to identify the semantic relationships between the entities. This also offers us an opportunity to solve other higher level NLP tasks, such as question answering, knowledge graph completion and information extraction. 
In the clinical domain, an entity can be a medical problem, a treatment, a medical test, body measurements etc. and the relationship between these entities can be diverse which are very much important to understand the clinical phenotypes.
\begin{exmp}
A biopsy of this mass was consistent with hematoma .
\label{exam2}
\end{exmp}
In the EMR sentence given in \ref{exam2}, the term ``a biopsy'' refers to a medical test. The terms ``this mass'' and the ``hematoma'' correspond to the entities related to medical problems. The aim of this task is to identify the underlying relationship that actually binds these together. %that is by which medical relationship they are bound together.
%The relation extraction task can be possible only when the entities in the text are identified. 
In 2010, the Informatics for Integrating Biology and the Bedside (i2b2) organized fourth i2b2/VA challenge \cite{rink2011automatic} which aims to extract $3$ coarse clinical relations (identifying the possible relations between medical problems and treatments, between medical problems and tests, and between pairs of medical problems) and other $11$ fine-grained relationships between the entities. \\
In recent past, this benchmark dataset has gained a lot of attention in biomedical NLP community, where several techniques have been proposed to solve this problem ranging from the semantic-injected kernel model to the machine learning embedded models. In recent past, with the success of deep neural network in various biomedical tasks \cite{kumar2016recurrent,yadav2018multi, yadav2016deep, ekbal2016deep, yadav2017entity}, attention of the researchers have shifted %the state-of-the-art models for this task has been drifted 
towards building relation extraction systems using deep learning frameworks. 
Some of the prominent methods include the works presented in \citet{de2011machine} which utilizes the kernel-based model to map features onto higher-dimensional space, \citet{sahu2016relation} that explored the convolutional neural network (CNN) driven technique extracting the relations at the sentence level etc. 
Inspired by \citet{doi:10.1093/jamia/ocx090}, \citet{yadav2018feature}, \citet{yadav2019feature} we focus on extracting the Shortest Dependency Path (SDP) between the entities, which helps in eliminating those words which are semantically not related. This enables our model to remove noise which hinders the performance of the system. Towards that, firstly we develop a dependency parser that determines the grammatical relationship between the words in the sentence. Details of the parser are provided in Section \ref{subsection:parser}. Given the success of the neural network, in this paper, we utilize Long Short-Term Memory (LSTM) \cite{hochreiter1997long} network, which has theoretically proven to cover the long-term dependencies. The proposed model takes as an input the embedding of the SDP based words which are assisted by the other latent features such as dependency labels, Part-of-Speech (PoS) information and the entities types. 
We use the benchmark i2b2/VA challenge dataset to validate our proposed model and for the easy comparison with the existing state-of-the-art techniques. Evaluation shows that our proposed method obtains significant improvements of F-score for most of the relation types except PIP. 
%\subsection{Contribution}
In summary the key contributions of our proposed work are as follows:\\
\textbf{1.} We develop a robust parser specific to the clinical texts. Its performance is found to be very close to the other state-of-the-art parsers.\\
\textbf{2.} We propose a shortest dependency path based LSTM model that provides state-of-the-art performance for relation extraction.
%\textbf{3.} Evaluation of the proposed work on the benchmark i2b2/VA 2010 challenge dataset.
\section{Related Works}
We consider the relation extraction as a multi-class classification task. Literature shows that most of the existing works in biomedical NLP are based on machine learning \cite{yadav2017feature,yadav2018information,yadav2017entity}.
A large body of works has been dedicated towards building robust relation extraction engines using traditional supervised machine learning models \cite{abacha2015text,singhal2016text}. %There are quite a large number of works that focused on extracting %Many models have been proposed to extract the 
%semantic relationship between the clinical concepts. 
A Support Vector Machine (SVM) was proposed in \newcite{rink2011automatic} for relation extraction. %based classifier . 
 The features used here was grouped into six classes: context features, similarity features, nested relation features, single concept features, Wikipedia features, and concept vicinity features. \newcite{minard2011multi} also used SVM for extracting the relations from the clinical reports. The features used in this task include the lists of medical abbreviations, features to capture the text writing style and semantic types of Unified Medical Language System (UMLS). %UMLS is widely used in information extraction in clinical and biomedical domains. 
\newcite{uzuner2010semantic} developed a semantic relation classification system based on SVM. They used UMLS to define medical problems, tests, and treatments. In recent years, deep neural network based methods are being widely used because of its many-fold benefits. %It does not require manual feature engineering. % to provide the state-of- the-art results. 
The advantage of using deep learning technique is its ability for extracting the features automatically. \newcite{sahu2016relation} relation extraction system employed Convolutional Neural Network(CNN) to learn features automatically and reduce the burden of manual feature extraction. To extract the relation between the entities in a sentence, it is not necessary to consider the whole sentence-rather only the shortest dependency path (SDP) between the entities as input to the system could achieve the robust state-of-the-art accuracy. \newcite{xu2015semantic} designed a semantic relation extraction system based on SDP. This showed that using a simple negative sampling technique can improve the performance of the system. 

Recurrent Neural Network (RNN) is used for sequence prediction problem like named entity recognition. But, when the input is very long then RNN suffers from vanishing gradient and exploding gradient \cite{hochreiter1997long}. To avoid these problems, LSTM (another variant of RNN) is widely used. \newcite{liu2017entity} used LSTM to identify entities from clinical text. The LSTM was used to extract the context information from the word representation sequence of the sentence.

Instead of relying on the full sentence, our proposed model considers only the SDP between the target entities as the input. It helps the model to eliminate those words which are not semantically important for deciding the relationship between the entities. Apart from the words in the SDP, we also consider other features like the POS tag information, the dependency label information, and the entity types. We use LSTM to extract the contextual information from the input features. The information generated by the LSTM is passed into a softmax layer to classify the type of relation.

%({\bf comment: with a new paragraph you should mention how is the current approach different from the other existing methods? })

\section{Proposed Methodology}
\begin{figure}[H]
  \scriptsize
  \includegraphics[width=9.5cm, height = 9cm]{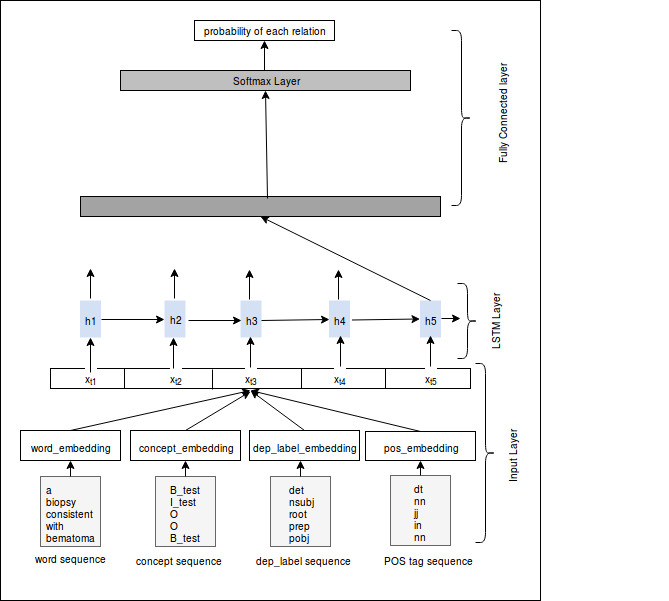}
  \caption{Our proposed architecture}
  \label{fig:architecture}
\end{figure}

Our proposed method for relation extraction is based on LSTM network. At first we develop a dependency parser specific to the clinical text, extract shortest dependency path between the clinical entities using developed parser, and finally use these information for determining the relations between the entities.\\ %In this section we present our proposed methodology. 
We ignore the sentences that do not contain at least two entities. %We consider only those sentences having at least two entities. 
This was done as the relation can exist only between the two entities. These sentence are then passed to the dependency parser that we develop. %Then the sentences are passed into a dependency parser. 
%({\bf comment: at first we should show here }
In the succeeding section, we provide detailed description of each steps. 
Overall architectural diagram is depicted in Figure \ref{fig:architecture}. 

%({\bf comment: we should have an overall architectural diagram first})
\subsection{Dependency Parser}\label{subsection:parser}
We develop our own dependency parser that is trained on the clinical text. The data used for training the parser is available as the constituency treebank. We convert the constituency treebank into dependency treebank using the Stanford CoreNLP \cite{de2006generating, manning2014stanford}. The output dependency trees are in the CoNLL-U format\footnotemark
\footnotetext{ \url{http://universaldependencies.org/format.html} }.
Our parser is based on the transition based algorithm developed by \newcite{nivre2003efficient}. In a transition based system, we convert the gold tree into the labeled configurations. The configurations of the parser are fed into the neural network, and train a model to predict the next transition for the unseen data. The configuration consists of a buffer which stores the words in the sentence which have not been processed yet, a stack which stores the words which have been processed but not yet assigned the head and a set which consists of the dependency labels. From the configuration, the top four words from the stack and their modifiers and first four front words in the buffer, as well as their corresponding PoS tags, are used as the features for our model. %{\bf comment: the previous sentence is incomplete. Re-write this}
There are variations of the transition based system. 
We follow the arc standard one for our parser. In this system, the initial configuration of the parser is given as, the stack contains a `Root' word, the buffer contains all the words of the sentences and the dependency label set is empty. This system defines three transitions.
  Say, S be the stack, B is the buffer, A be the set of the relations and i and j are the words in the sentences. The transitions are defined as follows:
 \begin{enumerate}
 \item LEFT-ARC(L):\\
    $\{S|i,j; B; A\} => \{S|j; B; AU\{jLi\}\}$; if the second top word in the stack is dependent on the top word of the stack then, a LEFT-ARC(L) operation is performed. Here, we pop the second top word from the stack and add the relation L between the top and second top element in the stack to the dependency relation set.
 \item RIGHT-ARC(L):\\
    $\{S|i,j; B;  A\} => \{S|i; B; AU\{iLj\}\}$; if the top word is dependent on the second top word of the stack then, a RIGHT-ARC(L) operation is performed. Here, we pop the top word from the stack and add the relation L between the top and second top element in the stack to the dependency relation set.
 \item SHIFT:\\
    $\{S|i; B|j;  A\} => \{S|i,j; B; A\}$; this operation simply sifts the first word from the buffer to the stack when there is no relation between the top two words in the stack.
 \end{enumerate}
To parse a new sentence, we use Genia tagger to get the PoS tags of the tokens\footnotemark \footnotetext{ \url{http://www.nactem.ac.uk/GENIA/tagger/} }. Then, we initialize the parser configuration. The parser provides the next transition by considering this configuration as the input. It will, thus, generate a dependency tree. We evaluate the performance of the parser with the labeled dependency treebank. This labeled dependency treebank is converted from manually annotated constituency treebank. The treebank consists of around fifty-three thousand sentences of clinical text which is provided by an industry named ezDI \footnotemark \footnotetext{ \url{https://www.ezdi.com/} }. %({\bf comment: labeled data of what? which data? details have to be mentioned here}) 
The result shows that our parser provides the state-of-the-art accuracies with 93.15 UAS and 92.01 LAS. Where, UAS is the ratio of the count of correctly parsed head position to the total number of the token and LAS is the ratio of the count of correctly parsed head position as well as correctly predicted dependency label to the total number of the token.

\subsection{Shortest Dependency Path-SDP}
The Shortest Dependency Path (SDP) is the shortest path between the two medical concepts in the dependency tree. For a pair of concepts (c1, c2) in a sentence, we find the shortest path by considering c1 as source vertex and c2 as the target vertex. Though the dependency tree of a sentence is a directed graph, we convert it into an undirected graph when we compute the SDP. 
\begin{figure}[!ht]
  \includegraphics[width=.95\linewidth]{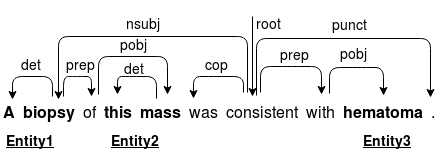}
  \caption{The dependency tree of the sentence given in Example~\ref{exam2}}
  \label{fig:dep_tree}
\end{figure}
In the dependency tree given in the Figure \ref{fig:dep_tree} of the sentence given in  Example~\ref{exam2}, the texts shown in the bold are the entities present in the sentence. The relations present in the sentence are TeCP (a biopsy, this mass), TeCP (a biopsy, hematoma) and PIP (hematoma, this mass).
%\begin{figure}[!ht]
  %\includegraphics[width=.95\linewidth]{img/SDP.png}
  %\caption{The SDP between \textbf{a biopsy} and \textbf{hematoma}}
  %\label{fig:sdp}
%\end{figure}
Considering the two entities \textbf{biopsy} and \textbf{hematoma}, the SDP between the two entities is generated as \textit{a biopsy consistent with hematoma}.
\\
 %Figure \ref{fig:sdp} shows the SDP between the two entities. 
The SDP is marked with concept type as \textit{B\_Test I\_Test O O B\_Problem} \footnotemark \footnotetext{Here B, I and O denote the beginning, intermediate and outside of an entity}. We term this sequence as a concept sequence. Now, the sequence of words in the SDP along with the concept sequence, the dependency labels sequence, and the PoS tags sequence are fed into the neural network. We use LSTM to learn the characteristic of the inputs and their contextual information. The model then outputs the probability values corresponding to all these relation types.
 
\subsection{Neural Network Architecture for Relation Extraction}
Our proposed architecture which is based on Long Short-Term Memory (LSTM) is composed of an input layer that takes SDP based words, an LSTM layer, and a softmax layer. 
We depict the proposed architecture in Figure \ref{fig:architecture}.
\subsubsection{Input layer}
The input layer has an embedding layer which maps each feature into the d-dimensional vectors. The vectors are then concatenated and passed into the LSTM layer. The feature vector of the t\textsuperscript{th} input is defined as
\[x\textsubscript{t} = E(w\textsubscript{t}) \oplus E(e\textsubscript{t}) \oplus E(d\textsubscript{t}) \oplus E(p\textsubscript{t}) \]
where w\textsubscript{t}, e\textsubscript{t}, d\textsubscript{t} and p\textsubscript{t} are the t\textsuperscript{th} word, concept tag, dependency label and PoS tag in the SDP, respectively. E maps the features into the vectors of dimension d and $\oplus$ is the concatenation operation. 

\subsubsection{LSTM Layer}
This layer takes the output of the embedding layer and computes the characteristic and contextual information of the input. The LSTM networks are a special kind of RNN. Unlike, the traditional neural network, the RNN cell at time t takes the input x\textsubscript{t} and previous hidden state h\textsubscript{t-1} to compute the current hidden state h\textsubscript{t}. But, RNN is not able to store the information of the long past inputs \cite{graves2013generating}. LSTMs are designed to overcome this issue. The key component in the LSTM is the cell state which carries information throughout the LSTM cells. An LSTM cell is composed of multiple components that control the information flow in the cell state. 

The first component is responsible for deciding which information to throw out from the cell state C\textsubscript{t-1}. It is controlled by a sigmoid layer called forget gate. It takes the previous hidden state h\textsubscript{t-1} and the current input x\textsubscript{t} and outputs a value between 0 - 1. Where, 1 stands for completely keeping the information in the cell state and 0 stands for completely ignoring the information. The forget gate is defined by the following equation. 
\[f\textsubscript{t} = \sigma (W\textsubscript{f}\cdot\big[ h\textsubscript{t-1},x\textsubscript{t}\big] + b\textsubscript{f})\]
Where, W\textsubscript{f} and b\textsubscript{f} are the weight matrix and the bias vector of the forget gate.
Now, the updated information in the cell state is \[C\textsubscript{t-1}  = C\textsubscript{t-1} * f\textsubscript{t}\]

The second component is responsible for deciding what new information to add in the cell state. It is controlled by a sigmoid layer called input gate (i\textsubscript{t}) and tanh layer. The tanh layer computes a new candidate $\overline{C\textsubscript{t}}$ that has to be multiplied component wise with the output of the input gate. \[i\textsubscript{t} = \sigma (W\textsubscript{i}\cdot\big[ h\textsubscript{t-1},x\textsubscript{t}\big] + b\textsubscript{i})\]
\[ \overline{C\textsubscript{t}} = tanh(W\textsubscript{c}\cdot\big[ h\textsubscript{t-1},x\textsubscript{t}\big] + b\textsubscript{c})\]
Now, the new cell state value at the current LSTM cell is given by
\[C\textsubscript{t} = C\textsubscript{t-1} + i\textsubscript{i} * \overline{C\textsubscript{t}}\]

The last component computes the hidden information of the LSTM cell. It consists of a sigmoid layer called the output gate(o\textsubscript{t}) and a tanh layer.  
\[o\textsubscript{t} = \sigma (W\textsubscript{o}\cdot\big[ h\textsubscript{t-1},x\textsubscript{t}\big] + b\textsubscript{o})\]
\[ h\textsubscript{t} = tanh(C\textsubscript{t}) * o\textsubscript{t}\]
Let H = (h\textsubscript{1}, h\textsubscript{2}, ... h\textsubscript{n}) be the hidden states produced by the LSTM layer. Where n is the length of the SDP. Now, the output of the last hidden state is passed fully connected layer for classification. 

\subsubsection{Softmax layer}
Softmax layer is used to classify the information that has been learned from the previous layers into relations. This layer converts the output of the LSTM layer into fixed length vector x of size same as the number of the relation classes n. Then, it computes the probability for all relation classes.

\section{Datasets and Experimental Setup}
\subsection{Dataset}
We use the i2b2-2010 relation extraction challenge dataset. The dataset is collected from three different hospitals viz, Partners Healthcare, Beth Israel Deaconess Medical Center, and the University of Pittsburgh Medical Center. It consists of discharge-summary and progress notes of the patients and is manually annotated by medical practitioners. We download the dataset from the i2b2 website. As mentioned in \newcite{sahu2016relation}, we also get only 170 documents for training and 256 documents for testing. But, the actual dataset released during the challenge was 394 documents for training and 477 documents for testing. The training data consists of 5264 relations and the testing data consists of 9069 relations. The Table \ref{annotation} shows the sample annotation format in i2b2 2010 dataset.
\begin{table}[H]
\scriptsize
\resizebox{.48\textwidth}{!}{%
\begin{tabularx}{.43\textwidth}{|l|X|X|r}
 \hline
  Text & 75  He did have burst of atrial fibrillation and was started on a Amiodarone gtt .\\ \hline
  Concept & \makecell { c=\enquote{burst of atrial fibrillation} 75:3 75:6||t= \enquote{problem}\\c=\enquote{a amiodarone gtt} 75:11 75:13||t=\enquote{treatment}} \\ \hline
Relation & c=\enquote{a amiodarone gtt} 75:11 75:13||r=\enquote{TrAP}||c=\enquote{burst of atrial fibrillation} 75:3 75:6 \\ \hline
\end{tabularx}}
\caption{\label{annotation} Sample annotation format in i2b2-2010 dataset}
\end{table}
\noindent In the first row, the number \textit{75} indicates the line number of the sentence in the document. The dataset consists of 8 relations:
\begin{enumerate}
\item TrIP: 
Treatment improves a medical problem. Example:- \textit{TrIP(po Amidarone||treatment, further episodes of AFIB||problem)} in the sentence, \textit{He had no further episodes of AFIB while on po Amiodarone .} 

\item TrWP: 
Treatment worsens a medical problem. Example:-  \textit{TrWP(increased nebulizer treatments||Treatment, upper respiratory like infection||problem)} in the sentence, \textit{This is a 55-year-old female with multiple prior admissions for pneumonia , COPD , asthma exacerbation, over 3 weeks of upper respiratory like infection unremitting with increased nebulizer treatments at home .}

\item TrCP:
Treatment causes a medical problem. Example:- \textit{TrCP(Drugs||treatment, Known Allergies||problem)} in the sentence, \textit{Patient recorded as having No Known Allergies to Drugs}.

\item TrAP: 
Treatment is administered for a medical problem. Example:-\textit{TrAP(CABG||treatment, MI||problem)} in the sentence, \textit{Father with MI in 50 's and underwent CABG .}. 

\item TrNAP: 
Treatment is not administered for a medical problem. Example:- \textit{TrNAP(ointments||treatment, incisions||problem)} in the sentence, \textit{No creams , lotions , powders , or ointments to incisions}.

\item TeRP: 
Medical test reveals a medical problem. Example:- \textit{TeRP(CXR||test, left lower lobe atelectasis||problem)} in the sentence, \textit{CXR 10-30 : Left lower lobe atelectasis has partially cleared .} 

\item TeCP: 
Medical test is conducted to investigate a medical problem. Example:- \textit{TeCP(cath||test, abnormal ett||problem)} in the sentence, \textit{67 y/o male with worsening shortness of breath. Had abnormal ETT and referred for cath .} 

\item PIP:
Medical problem interacts with a medical problem. Example:- \textit{PIP(wounds||problem, infection||problem)} in the sentence, \textit{Monitor wounds for infection - redness , drainage , or increased pain}. 
\end{enumerate}
\subsection{Experimental Setup}
During preprocessing, we filter out all the sentences having entity lesser than two. %Then, for every pair of entities in a sentence, we get their relation type from the annotated corpus. If there is no relation between the entities, then we mark their relation as NONE. Then, the sentences having more than one entity are passed into a dependency parser. From the dependency parser output of the sentence, the SDP between every pair of entities present in the sentence is computed. The entities in the SDP are marked with their concept type. To define the boundary of the entity, the first words of the entity is marked as B\_concept type. And the remaining parts of the entity are marked with I\_concept type and non-entity words in the SDP are marked with O. Where B stands for beginning and I stands for intermediate and O stands for others. 
Since the number of instances is very low in the training set, we %combine the training data and testing data together and perform 
perform five-fold cross-validation on the combined dataset of training and test. As mentioned in the above section, we use four features, the word sequence, the concept sequence, the dependency label sequence and the PoS tag sequence of the SDP. For word embedding we use pre-trained word vector trained using word2vec tool \cite{mikolov2013distributed} on huge Pubmed articles. This word embedding is downloaded from \url{http://bio.nlplab.org/}. To generate embeddings for the remaining features i.e. POS tag, dependency label, and the entity type, we train our own vectors using gensim word2vec tool. %({\bf comment: the colored texts are not clear. Did you use word embeddings created from PubMed? For where did you download this? For what purpose you used gensim word vectors then?})
We set the dimension of the vectors to 50 because of the small vocabulary size. We consider the instances of all the relation types including the NONE class. We train two models, one based on our developed parser and the other based on the state-of-the-art Stanford parser.

\subsection{Hyperparameters}
We use categorical cross-entropy as loss function for our neural network and the rmsprop as the optimizer. The number of neurons in the LSTM layer is 512. The next hidden layer consists of 256 neurons. We use Relu as an activation function in this layer. The output layer consists of 9 neurons which correspond to relation types. Since the problem is a classification problem, we apply softmax activation function in the output layer. We use dropout value as 0.3 to overcome the overfitting problem. The network is trained with 50 epochs.

\section{Experimental Results}
All the results shown below are the average of 5-fold experiments. To show the effectiveness of our proposed model, we build two competitive state-of-the-art baseline models.
In the first baseline, we train the model with feed forward network and the second baseline model is similar to our proposed model but trained without dependency label and PoS tag features.\\
%\subsection{Comparison with baseline models}
From the Table \ref{basevsproposed}, we can see that our proposed model achieves an improvement around $2\%$ F1-score over the first baseline and around $0.56$ F1-score over the second baseline. 
\begin{table}[H]
\scriptsize
\begin{center}
 \begin{tabular}{c|c|c|c} 
 \hline
 \hline
 Model & Precision & recall & F1-score\\ [0.5ex] 
 \hline
 \hline
 Baseline model 1 &89.97  &90.86  &90.27\\

 Baseline model 2 &92.19   &92.43  &92.24\\ 

 Our proposed model &92.79  &92.87  &92.80\\

\end{tabular}
\end{center}
\caption{\label{basevsproposed} Result comparison of our baseline models and the proposed model}
\end{table}

\subsection{Comparisons with state-of-the-art Systems}
There are quite a few existing works which focused on experiments on the i2b2-2010 full dataset, but we found only \newcite{sahu2016relation} which performed the experiment on the i2b2-2010 partial dataset which is same as the dataset used in our experiment. We are not able to perform the experiments on the whole dataset because the whole dataset was only released during the i2b2-2010 challenge. It consists of 394 documents for training and 477 documents for testing \cite{rink2011automatic}. But when we download the dataset from the i2b2 website, we got only 170 documents for training and 256 documents for testing.
%({\bf comment: an obvious question is why we have not been able to perform experiment on the whole dataset?})
Results shown in Table \ref{sk_SahuVsOurmodel} demonstrates that our proposed model provides better performance for all the relation types except PIP. 
\begin{table}[H]
\scriptsize
\begin{center}
 \begin{tabular}{c|c|c} 
 \hline
 Relation Type & Proposed model & \newcite{sahu2016relation} model\\ [0.5ex] 
 \hline
 \hline
  TeCP &\textbf{59.13}  & 50.56\\
 
 TrCP  &\textbf{62.13} & 56.44 \\
 
 PIP & 59.38 & \textbf{64.92} \\

 TrAP  &\textbf{75.35} & 69.23 \\
 
 TeRP  &\textbf{83.86} & 81.25 \\
 
\end{tabular}
\end{center}
\caption{\label{sk_SahuVsOurmodel} Comparisons with \newcite{sahu2016relation}}
\end{table}

%\subsection{Comparison of our parser based model and the Stanford parser based model}

\begin{figure}[H]
  \includegraphics[width=7.8cm, height=5.5cm]{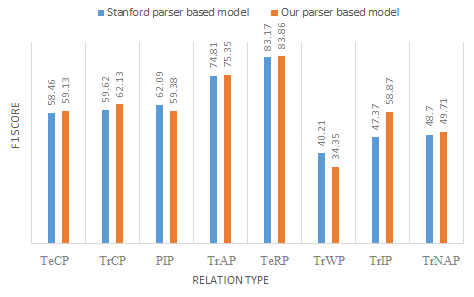}
  \caption{Comparison: Our parser based model vs. Stanford parser based model}
  \label{fig:re_com}
\end{figure}

In order to show how our proposed parser performs against the Stanford parser we develop two variations of our proposed relation extraction model, one by using the dependency relations extracted from our parser and the other by using the relations extracted from the Stanford parser. Performance as shown in Figure \ref{fig:re_com} shows that for most of the relation types, our parser based model outperforms the Stanford parser based model. Due to space constraint we can't include the result for NONE to the graph. For NONE, our parser based model attained an F-1 score of $96.50$ compared to $96.58$ F1-score of the Stanford parser.% based model provides an F-1 score of $96.58$.
Overall evaluation results of our parser based model and the Stanford parser based model are shown in Table \ref{ourPvsStf}.
\begin{table}[H]
\scriptsize
\begin{center}
 \begin{tabular}{|c|c|c|c|} 
 \hline
 Model & precision & recall & F1-score\\ [0.5ex] 
 \hline
 Our parser based model &92.79  &92.87  &92.80\\
 \hline
 Stanford parser based model &92.88  &92.92  &92.86\\ 
 \hline
\end{tabular}
\end{center}
\caption{\label{ourPvsStf} Overall evaluation: Our Parser based model vs. Stanford parser based model}
\end{table}
\subsection{Hypothesis Testing}
 We show in Table \ref{hypo} the results of hypothesis testing. 
 \begin{table}[H]
\scriptsize
 \resizebox{.48\textwidth}{!}{%
\begin{tabularx}{.53\textwidth}{l|X|X X }
 \hline
 Model & T-Value & P-Value \\ [0.5ex] 
 \hline
 \hline
BL1 &26.34559  & < .00001  \\

BL2 &4.22267  &.001453  \\ 

Stanford parser& -0.41611 &.344136 \\

\end{tabularx}}

\caption{\label{hypo} Significance testing w.r.t our proposed model. Here, BL1 and BL2 denote the first and the second baseline, respectively}
\end{table}

\noindent The first row, the second row, and the third row are the result of hypothesis testing of our proposed model against the baselines and % model 1, against the baseline model 2, and 
the Stanford parser based model, respectively. From the hypothesis testing, it is evident that performance improvement in our proposed model is statistically significant over the baselines. However, we observe that the performance improvement of the Stanford parser based model is not significant over our parser based model. Instead the performance our parser is very close to the Stanford parser based model.
\subsection{Error Analysis}
\noindent
We conduct detailed error analysis on our predicted outputs. 
%Next, we are comparing the predicted outputs of the two models and trying to find out the errors.
Below we provide an example for error analysis of the predicted output.
\begin{exmp}
\# Neurologic - The patient was seen by the Neurology consult service and underwent MRI head which revealed lesions suspicious for metastases, possible hemorrhages, and findings consistent with hypoxic brain injury.
\label{exam3}
\end{exmp}
In this example, the two parsers do not produce the identical outputs. The possible entity pairs present in the sentence are (lesions||Problem,  metastases||Problem), 
(mri head ||Test, hypoxic brain injury||Problem), 
(mri head||Test, hemorrhages||Problem), 
(mri head||Test, metastases||Problem), 
(mri head||Test, lesions||Problem), 
(lesions||Problem, hypoxic brain injury||Problem), 
(lesions||problem,  hemorage||Problem), 
(metastases||problem, hypoxic brain injury||Problem), 
(metastases||problem, hemorage||Problem), and 
(hemorage||Problem, hypoxic brain injury||Problem). 
\begin{table}[H]
\centering
\scriptsize
%\resizebox{width= 0.5\linewidth
\resizebox{.48\textwidth}{!}{%
\begin{tabularx}{.48\textwidth}{|r|X|X|X|l}
\hline
Actual relation & Our parser based model output & Stanford parser based model output     \\ \hline
 PIP &                          \textbf{NONE}                   &PIP \\ \hline
TeRP &                 TeRP                    & \textbf{NONE}\\ \hline
TeRP&                           TeRP                     &TeRP\\ \hline
 TeRP&                             TeRP                   &TeRP\\ \hline
TeRP&                               TeRP                 & TeRP\\ \hline
 NONE&           NONE                        & \textbf{PIP}\\ \hline
 NONE&                   NONE                           &NONE\\ \hline
 NONE&     NONE                            &NONE\\ \hline
NONE&              NONE                          & NONE\\ \hline
NONE&      NONE                           & NONE\\ \hline
\end{tabularx}}

\caption{Actual relations present in the sentence given in Example \ref{exam3} and corresponding outputs of our parser based model and the Stanford parser based model.}
\label{table:sent1:output}
\end{table}
\noindent Table \ref{table:sent1:output} shows all the correct relations present in the sentence given in Example \ref{exam3} and the corresponding outputs of our parser based model and the Stanford parser based model. From Table \ref{table:sent1:output}, we can see that two parser models produce different outputs for three relations. The possible reasons for the incorrect outputs are noted below:  
\begin{enumerate}
\item Actual relation is PIP(lesions||Problem,  metastases||Problem). Our parser based model yields the output NONE whereas the Standford parser based model's output is PIP. And the SDP generated by both the parsers is \enquote{lesions suspicious for metastases}. For this relation, the output of the Stanford parser is correct even if the SDP generated by these two parsers are same.
\item 
The actual relation is TeRP(mri head||Test, hypoxic brain injury||Problem). Our parser based model's prediction is TeRP and the SDP is \enquote{mri head revealed lesions suspicious for metastases findings consistent with hypoxic brain injury}. The Stanford parser based model's prediction is NONE and the SDP is \enquote{mri head underwent seen findings consistent with hypoxic brain injury}. In this case, our parser based model provides the correct relation. It may be the reason that our parser based SDP provides a better sequence of words. In the SDP generated by the Stanford parser, the word \enquote{findings} depends on the word \enquote{seen} which is incorrect because the words \enquote{hemorage} and \enquote{findings} should depend on the word \enquote{metastases} with \enquote{conj} relation. But our parser provides correct dependency tree.
\item
Actual relation is NONE(lesions||Problem,  hypoxic brain injury||Problem) Our parser based model's output is NONE. Our parser based SDP is \enquote{lesions suspicious for metastases findings consistent with hypoxic brain injury}. The Stanford parser based model's output is PIP. The Stanford parser based SDP is \enquote{lesions revealed head underwent seen findings consistent with hypoxic brain injury}. Here also, our parser based model provides a correct relation type but the Stanford parser based model provides the wrong relation. It may be the reason that our parser based SDP provides a better sequence of words.
\end{enumerate}
In the example \ref{exam2}, three entity pairs are present i.e. (a biopsy||test, this mass||problem), (a biopsy||test, hematoma||problem), and (hematoma||problem, this mass||problem). For this example, both the parser gives same dependency tree but the Stanford parser provides a better output. Both the models give a correct relation type for the first entity pairs i.e. TeCP and wrong relation type for the second entity pair i.e. TeRP instead of TeCP. But for the last entity pair, the Stanford parser based model provides correct relation type i.e PIP whereas our parser based model provides wrong relation type i.e. NONE.

%\begin{table}[H]
%\centering
%\scriptsize
%\resizebox{width= 0.5\linewidth
%\resizebox{.48\textwidth}{!}{%
%\begin{tabularx}{.48\textwidth}{|r|X|X|X|l}
%\hline
%Actual relation & Our parser based model output & Stanford parser based model output     \\ \hline
% TeCP &                         TeCP                  &TeCP\\ \hline
%TeCP &                 \textbf{TeRP}                    & \textbf{TeRP}\\ \hline
%PIP&                           \textbf{NONE}                     &PIP \\ \hline
%\end{tabularx}}

%\caption{Actual relations present in the sentence given in Example \ref{exam4} and corresponding outputs of our parser based model and the Stanford parser based model.}
%\label{table:sent2:output}
%\end{table}

%({\bf but the performance of Stanford parser is better than ours. So there would be counter examples where Stanford parser based approach could perform better. })

\section{Conclusion}
In this paper we propose an effective model for relation extraction in clinical text. At first we develop a parser for clinical domain, and then use the dependency relations extracted from the parser as a feature to the deep learning model for relation extraction. %With our simple approach, we could achieve a better result as compared to the existing model. Hence, 
We consider the SDP generated by the dependency parser a better feature representation for relation extraction. By comparing the results to the baseline models, we can conclude that LSTM based model can extract the contextual and the sequential information from the SDP and the dependency label and PoS tag information can enhance the performance of the model. Our experimental results also show that the overall performance of our parser based model is very close to the overall performance of the Standford parser based model. However, our detailed analysis reveals that our parser based model makes correct prediction for several instances for which Stanford parser based model makes wrong predictions. %correctly assigns attains better performance But there are instances for which But if we look into relation wise prediction, our parser based model provides a better result.

% include your own bib file like this:
%\bibliographystyle{acl}
%\bibliography{acl2018}
\bibliography{acl2018}
\bibliographystyle{acl_natbib}

\end{document}